\documentclass[conference]{IEEEtran}
\IEEEoverridecommandlockouts

\usepackage{amsmath,amssymb,amsfonts, booktabs}
\usepackage{algorithm}
\usepackage{algpseudocode}
\usepackage{graphicx}
\usepackage{textcomp}
\usepackage{subcaption}
\usepackage{multirow}
\usepackage[hidelinks]{hyperref}

\usepackage{tikz}
\usepackage{tikz-layers}
    \usetikzlibrary{arrows,calc,fit,matrix,patterns,positioning,shapes,shadows}
\usepackage[style=ieee,isbn=false,doi=false,backend=biber,maxnames=1,citestyle=numeric-comp]{biblatex}
\usepackage{todonotes}

\usepackage{ifthen}
\newboolean{generateTikzfigures}
\setboolean{generateTikzfigures}{false}
\newcommand\ext{pdf}

\newsavebox\SparsePlannerTeaser

\ifthenelse{\boolean{generateTikzfigures}}{
    \usepackage{pgfplots}
    \usepackage{tikz}
    \usepackage{tikz-layers}
    \usetikzlibrary{calc,fit,positioning}
    \pgfplotsset{compat=1.18}
    \usepgfplotslibrary{external}
    \tikzexternalize[prefix=./images/,optimize=false]
}{}

\algblock{ParFor}{EndParFor}
\algnewcommand\algorithmicparfor{\textbf{parfor}}
\algnewcommand\algorithmicpardo{\textbf{do}}
\algnewcommand\algorithmicendparfor{\textbf{end\ parfor}}
\algrenewtext{ParFor}[1]{\algorithmicparfor\ #1\ \algorithmicpardo}
\algrenewtext{EndParFor}{\algorithmicendparfor}

\newcounter{algoline}[algorithm]

\makeatletter
\patchcmd{\ALG@step}{\addtocounter{ALG@line}{1}}{\stepcounter{ALG@line}\refstepcounter{algoline}}{}{}

\makeatother

\makeatletter
\apptocmd{\@maketitle}{\centering\insertfig}{}{}
\makeatother

\newcommand{\insertfig}{%
    \setlength{\fboxsep}{0pt}%
    \resizebox{\textwidth}{!}{
        \ifthenelse{\boolean{generateTikzfigures}}{
            \usebox{\SparsePlannerTeaser}
        }{
            \includegraphics[width=\textwidth]{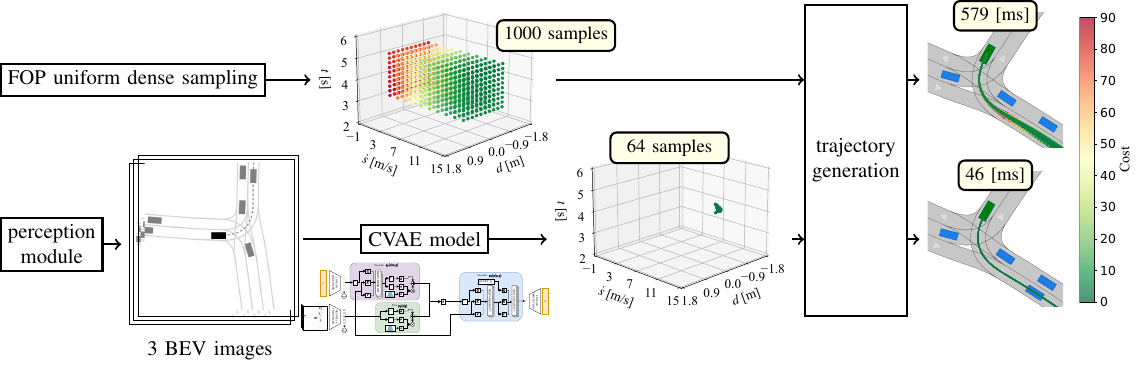}
        }
    }
\setcounter{figure}{0}
\captionof{figure}{The scheme illustrates the idea of the proposed Sparse Planner. The uniform dense sampling in the Fren\'{e}t parameter space is replaced with a CVAE-based sampling model, which learns an efficient sampling distribution over $(d, \dot{s}, t)$ conditioned on the Bird’s Eye View (BEV) representation, while leaving the downstream trajectory generation module unchanged.
}
\label{fig:scheme}}

\begin{document}

\ifthenelse{\boolean{generateTikzfigures}}{
\tikzsetnextfilename{teaser}
\begin{lrbox}{\SparsePlannerTeaser}
    \begin{tikzpicture}[box/.style={rectangle,draw},
                    draw=black,node distance=1cm,line width=0.1em,font=\linespread{1.0}\selectfont]
        \node[] (costmap) {\includegraphics[height=5cm]{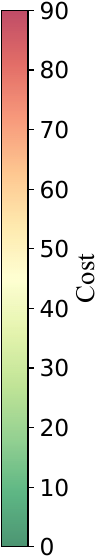}};
        \node[left=0cm of costmap.south west,anchor=south east] (sparseplanner) {\includegraphics[height=2.3cm]{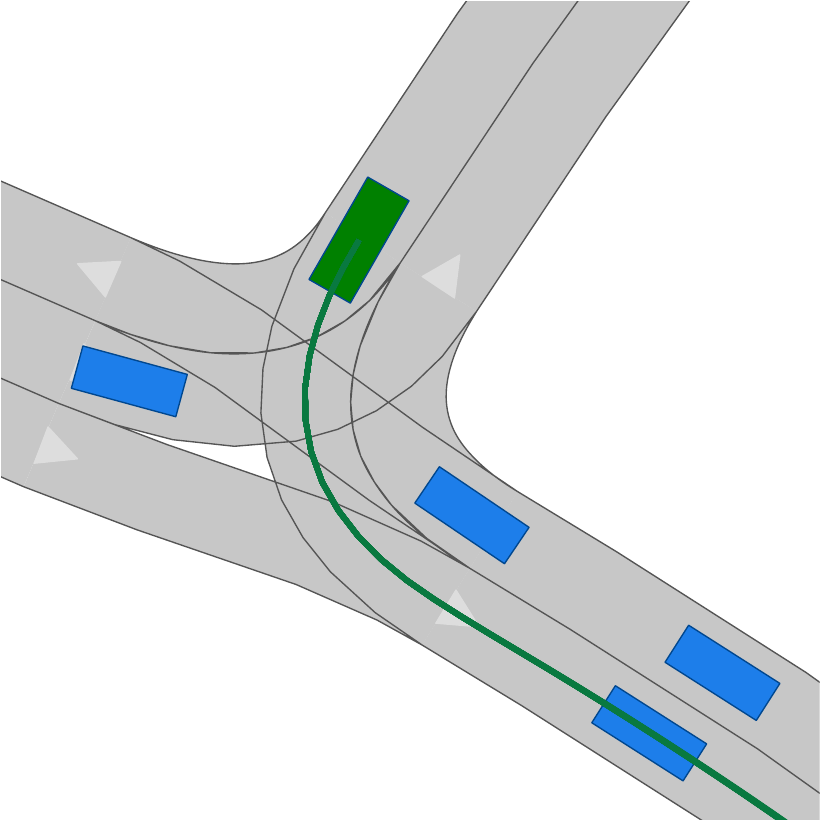}};
        \node[left=0cm of costmap.north west,anchor=north east] (fop) {\includegraphics[height=2.3cm]{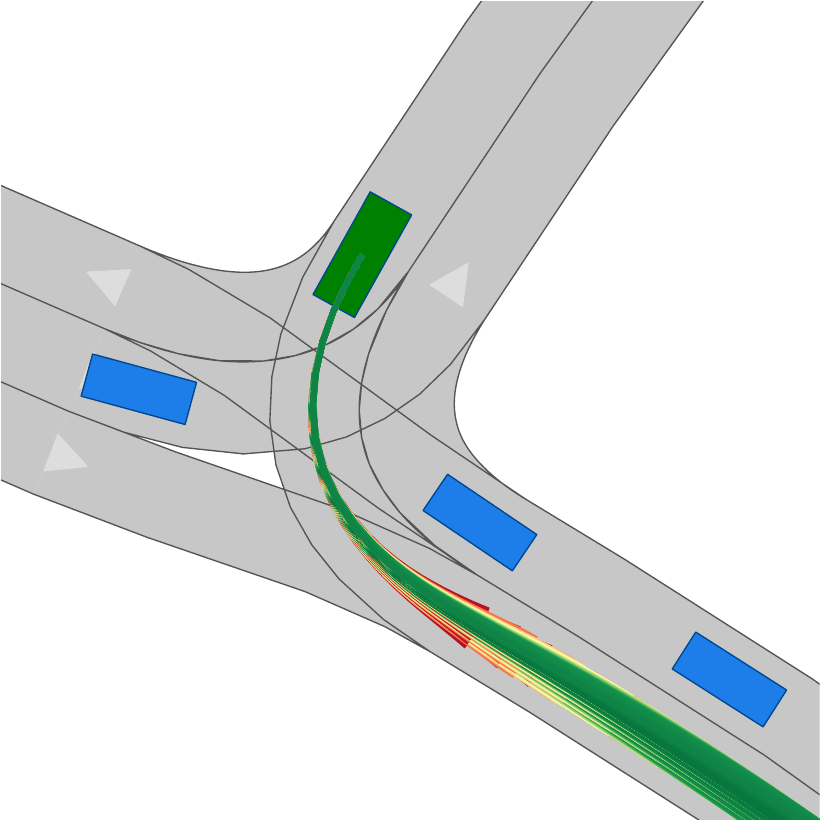}};
        \path let \p1=($(fop.north)-(sparseplanner.south)$),\n1={veclen(\p1)} in node[box,minimum height=\n1,left=0.2cm of fop.north west,anchor=north east,align=center] (tragen) {trajectory\\generation};
        \node[left=0.2cm of tragen.south west,anchor=south east] (path1) {\includegraphics[height=2.8cm]{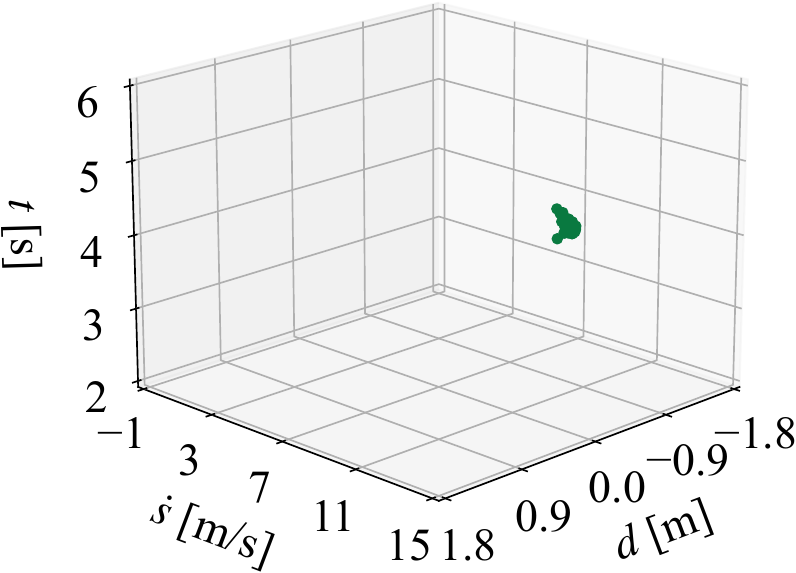}};
        \node[left=4.2cm of tragen.north west,anchor=north east] (path2) {\includegraphics[height=2.8cm]{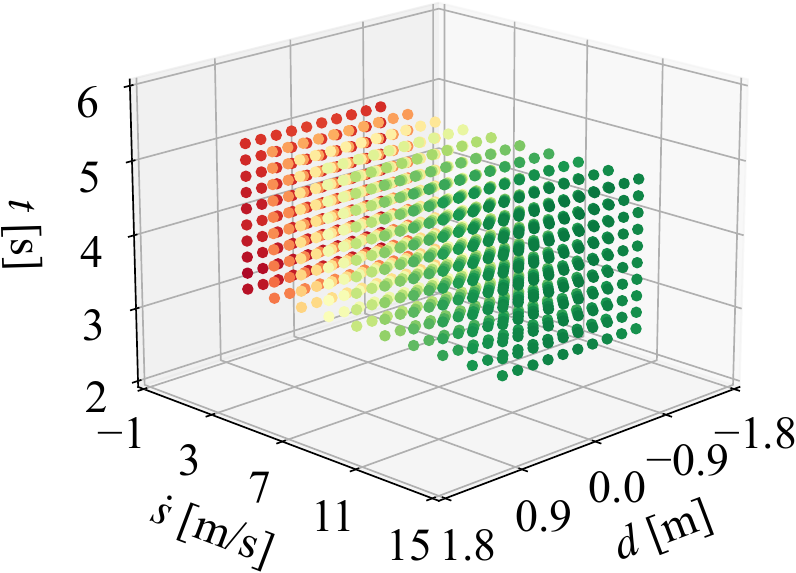}};
        \node[below left=1cm and 0.2cm of path2.south west,anchor=east,label=below:{3 BEV images}] (BEV1) {\fbox{\includegraphics[height=2.5cm]{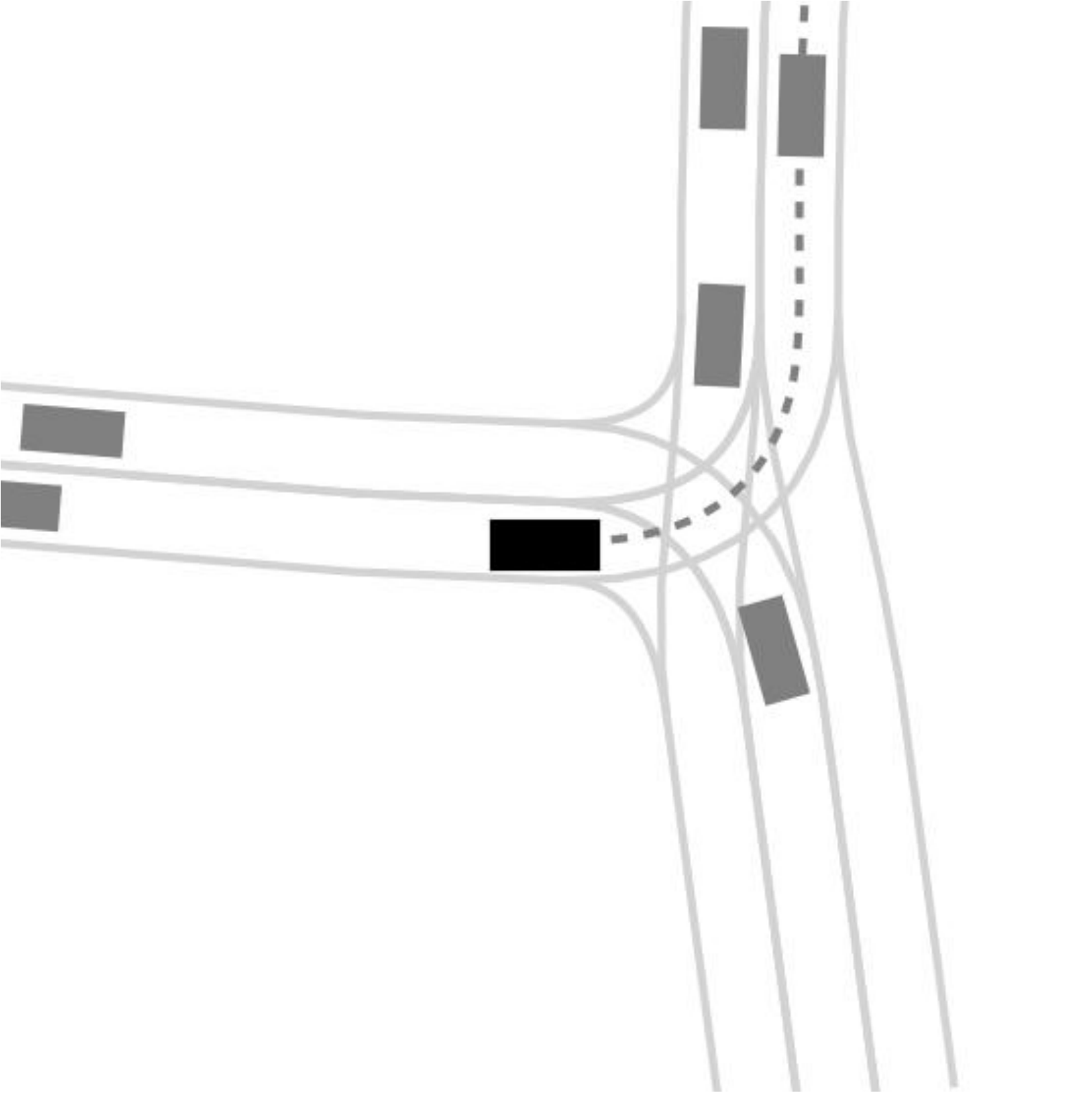}}};
        \node[above right=0.1cm of BEV1.south west,anchor=south west] (BEV2) {\fbox{\includegraphics[height=2.5cm]{images/scheme_images/BEV}}};
        \node[above right=0.1cm of BEV2.south west,anchor=south west] (BEV3) {\fbox{\includegraphics[height=2.5cm]{images/scheme_images/BEV}}};
        \node[box,left=0.3cm of BEV1,align=center] (pm) {perception\\module};
        \node[box,anchor=west,align=center] (ds) at (pm.west |- fop) {FOP uniform dense sampling};
        \draw[->] (path1.east |- sparseplanner) -- (tragen.west |- sparseplanner);
        \draw[->] (path2.east |- fop) -- (tragen.west |- fop);
        \draw[->] (tragen.east |- sparseplanner) -- (sparseplanner);
        \draw[->] (tragen.east |- fop) -- (fop);
        \draw[->] (ds) -- (path2.west |- ds);
        \draw[->] (pm) -- (BEV1.west |- pm);
        \draw[->] (BEV2.east |- sparseplanner) -- node[box,midway,fill=white] (cvae) {CVAE model} (path1.west |- sparseplanner);
        \node[below=0cm of cvae] {\hyperref[fig:architectures]{\includegraphics[width=4.2cm]{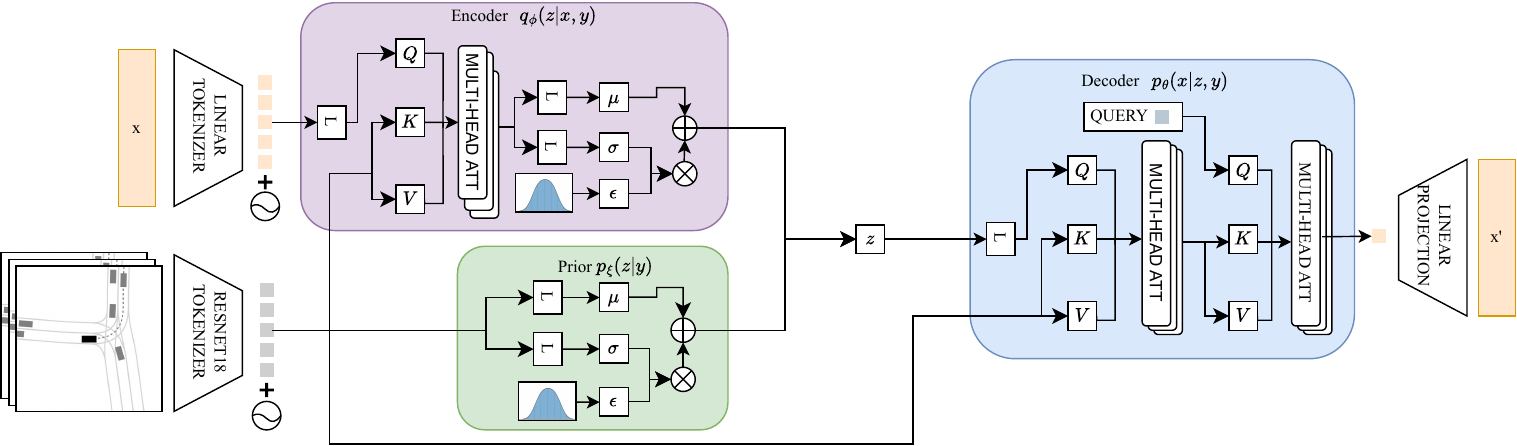}}};
        \node[rectangle,draw,fill=yellow!10,rounded corners=3pt,minimum width=2cm,align=center,font=\small] at ($(path2.east)+(0,1cm)$) {1000 samples};
        \node[rectangle,draw,fill=yellow!10,rounded corners=3pt,minimum width=2cm,align=center,font=\small] at ($(path1.north)-(0,0.2cm)$) {64 samples};
        \node[rectangle,draw,fill=yellow!10,rounded corners=3pt,minimum width=1cm,align=center,font=\small] at ($(fop.north)-(0,0.2cm)$) {579~[ms]};
        \node[rectangle,draw,fill=yellow!10,rounded corners=3pt,minimum width=1cm,align=center,font=\small] at ($(sparseplanner.north)-(0,0.2cm)$) {46~[ms]};
    \end{tikzpicture}
\end{lrbox}
}{}

\title{Sparse Planner: A Hybrid Planner for Efficient Sampling via a Conditional Variational Autoencoder}

\author{
\IEEEauthorblockN{
Wenguang Xu$^{1}$,
Giovanni Lucente$^{2}$,
Karem Mohamed$^{1}$,
Richard Membarth$^{1,3}$
}

\IEEEauthorblockA{
wenguang.xu@thi.de,\;
giovanni.lucente@dlr.de,\;
karem.mohamed@thi.de,\;
richard.membarth@thi.de
}

\thanks{*This work is supported by the Federal Ministry of Education and Research (BMBF) as part of the MANNHEIM-AutoDevSafeOps project, by the Bavarian State Ministry of Science and the Arts (StMWK) as part of the FlexType project as well as by the Federal Ministry for Economic Affairs and Energy (BMWE) as part of the nxtAIM project.}
 \thanks{$^{1}$Research Institute AImotion Bavaria, Technische Hochschule Ingolstadt (THI), Ingolstadt, Germany}
\thanks{$^{2}$Research Institute of Transportation Systems, German Aerospace Center (DLR), Braunschweig, Germany} \\
\thanks{$^{3}$German Research Center for Artificial Intelligence (DFKI), Saarbrücken, Germany}
}

\maketitle

\begin{abstract}
Trajectory planning is a core component of autonomous driving systems, where real-time performance and solution quality directly affect safety and reliability. Sample-Based Motion Planning (SBMP) is widely adopted for its ability to approximate near-optimal solutions through parameter space sampling. However, achieving high-quality trajectories typically requires dense sampling, leading to substantial computational overhead and significant runtime variability in complex traffic scenarios.

To address this limitation, we propose a Sparse Planner (SP) that improves sampling efficiency by learning the conditional relationship between scene context and effective trajectory parameters using a Conditional Variational Autoencoder (CVAE). By modeling the structure of high-quality sampling distributions, SP directly generates cost-effective samples in the parameter space, significantly reducing the required sampling density while preserving solution quality.

Experimental results show that SP achieves lower trajectory cost than the state-of-the-art FISS+ planner while using only one-eighth of the sampling density. In addition, SP demonstrates improved distance-keeping capability in obstacle-rich scenarios and maintains reduced and more stable runtime characteristics, indicating enhanced computational efficiency and predictable runtime behavior.
\end{abstract}


\section{Introduction}
Trajectory planning is a core module in autonomous driving systems, where real-time performance directly impacts driving safety.
With the increasing complexity of autonomous driving functionalities, the computational consumption of planning modules continues to grow.
Improving runtime efficiency while maintaining functional reliability and real-time performance has therefore become a critical challenge.
This practical demand constitutes the primary motivation for this work.

In recent years, the rapid development of deep learning has motivated extensive research into data-driven approaches that directly predict control commands or complete trajectories from perception or environmental representations, serving as alternatives to conventional physics-based planners.
However, such end-to-end learning-based methods often suffer from limited generalization in long-tail scenarios, raising concerns regarding safety and robustness.
Moreover, the lack of interpretability in their decision-making process remains a fundamental obstacle to reliable deployment.
More recently, the emergence of large language models (LLMs) has stimulated renewed interest in hybrid autonomous driving architectures, in which learning-based models are employed for high-level semantic understanding and decision reasoning, while low-level trajectory generation is still handled by conventional rule--based planning algorithms~\cite{Xu2025ChatMPC, wang2024dualad, PlanAgent2026, Wang2023DriveMLMAM}.
In this context, SBMP methods regain prominence due to their explicit constraint handling, probabilistic completeness, and scalability.

However, conventional SBMP relies on dense sampling over a parameter space to approximate an optimal solution within a discretized candidate set, which leads to a severe combinatorial explosion and limits real-time applicability.
To address this issue, we propose a learning-assisted sampling-based planner, termed Sparse Planner (SP).
By leveraging a conditional variational autoencoder (CVAE), SP transforms conventional dense uniform sampling into a sparse, cost-biased sampling process, significantly reducing the number of required samples while preserving trajectory quality.
As a byproduct of the CVAE-based sampling, the learned distribution implicitly concentrates the samples in collision-free regions of the trajectory space.
The proposed SP runs at an average of 47\,ms per planning step, of which 7\,ms is spent on CVAE inference.

This paper makes the following contributions:
\begin{itemize}
    \item proposes SP, a hybrid planner based on a CVAE that achieves lower trajectory cost while maintaining low, bounded, and predictable runtime behavior (\autoref{sec:methodology});
    \item introduces a CVAE architecture featuring new design elements and a tailored loss formulation to improve sampling effectiveness (\autoref{sec:CVAE_optimization}).
    \item reduces the required sampling number to one-eighth and achieves a $2.5\times$ speedup over the state-of-the-art (FISS+) for comparable trajectory cost levels, while also decreasing collision-induced planning failures.(\autoref{sec:evaluation}).
\end{itemize}

\section{Background}
\label{sec:background}
Trajectory planning over the horizon $[0,T]$ can be formulated as a constrained optimization problem:
\begin{equation}
    \label{eq:planning_problem}
    \begin{aligned}
             \tau^{*} &= \arg\min_{\tau \in \mathcal{T}} J(\tau) \\
    \text{s.t.} \quad & \tau(0) = x_{\text{init}}, \quad \tau(T) \in \mathcal{X}_{\text{goal}}, \\
                      & \tau(t) \in \mathcal{X}_{\text{free}}(t), \quad \forall t \in [0,T], \\
                      & g(\tau(t)) \le 0, \quad \forall t \in [0,T].
    \end{aligned}
\end{equation}
where $\mathcal{T}$ denotes the continuous trajectory space and $J(\tau)$ represents the trajectory cost.
The trajectory starts from the initial vehicle state $x_{\text{init}}$, terminates within the goal region $\mathcal{X}_{\text{goal}}$, and remains inside the time-varying collision-free state space $\mathcal{X}_{\text{free}}(t)$.
The inequality constraint $g(\tau(t)) \le 0$ captures vehicle dynamic limits, passenger comfort requirements, and traffic regulations.
Trajectory planning algorithms can generally be categorized into graph-, sampling-, optimization-, and learning--based methods.

Graph-based planners such as Dijkstra~\cite{Dijkstra} and A*~\cite{Hart1968A} discretize the state space and search for feasible paths over the resulting graph.
These methods typically provide completeness guarantees and are used for global path planning tasks.
However, due to the inherently discrete representation of paths, vehicle dynamics, and temporal continuity are often modeled in a simplified or indirect manner, which makes it difficult to generate continuous, dynamically feasible trajectories.

Optimization-based trajectory planners, such as DeepGame-TP~\cite{DeepGameTP} and Bertha~\cite{Ziegler2014Bertha}, formulate trajectory generation as an optimization problem, in which a cost function is minimized over a finite discretized time horizon with respect to the  control variables, while explicitly enforcing kinematic feasibility and safety constraints.
Although fine-grained parallelism \cite{xu2026trajectory} can be exploited for inner loop kernel using techniques such as single instruction multiple data (SIMD), the iterative nature of these methods introduces strong data dependencies, which fundamentally limit algorithm-level parallelization.
Moreover, their runtime behavior is often difficult to predict, posing challenges for achieving stable real-time performance in complex scenarios or on resource-constrained platforms.

In recent years, learning-based trajectory planning methods, such as UniAD~\cite{hu2023_uniad} and BridgeAD~\cite{Zhang2025Bridging}, have emerged as data-driven alternatives that map environmental states or perception features into a compact latent representation, enabling direct prediction of control commands or trajectories with low inference latency at runtime.
However, these methods typically lack explicit modeling of vehicle dynamics and safety constraints, resulting in limited interpretability and safety guarantees in safety-critical applications~\cite{Chen2024Challenges}.

In contrast, sampling-based planners approximate the continuous planning problem by enumerating a finite set of dynamically feasible motion candidates.
A representative method in this category is the Fren\'{e}t Optimal Planner (FOP) \cite{Werling2010FOP}, which specializes sampling-based planning by using a low-dimensional Fren\'{e}t coordinate representation.
The Fren\'{e}t frame is defined based on a reference path given from the global planner, so that lateral and longitudinal motions can be decoupled and parameterized independently.
FOP generates candidate trajectories by sampling a small set of interpretable terminal parameters over a planning horizon $T$, where the lateral motion is specified by a terminal offset $d$ and the longitudinal motion by a target velocity $\dot{s}$.
Smooth trajectories are constructed in the Fren\'{e}t frame using polynomial functions and subsequently transformed into the global coordinate system.
Feasibility checks are then performed to ensure compliance with vehicle dynamics and safety constraints, that is, $g(\tau(t)) \le 0$ and $\tau(t) \in \mathcal{X}_{\text{free}}(t)$.
Finally, a cost function is evaluated for each feasible candidate, and the trajectory with minimum cost is selected.
Since candidate trajectories are generated and evaluated independently, FOP exhibits no inter-trajectory data dependency, making it particularly suitable for parallel computation.

\section{Related Work}
\label{sec:related_work}

\subsection{Heuristics-Driven Sampling}
To address the trade-offs between computational efficiency and trajectory optimality, recent research has introduced iterative search strategies that avoid exhaustive uniform sampling.
The Fast Iterative Search and Sampling (FISS) planner~\cite{Sun_FISS_2022} pre-ranks candidate samples using a priority queue based on estimated costs.
These estimates rely on a history-based heuristic that biases the search toward the previous cycle’s optimal solution.
Building upon this, FISS+~\cite{Sun2023FISSPLUS} utilizes a two-stage coarse-to-fine framework that combines a discretized global search with a refinement stage using gradient descent.
The refinement stage improves the solution in the continuous space, reducing discretization-induced suboptimality and enables a more accurate approximation of the global optimum.
Due to its strong performance, FISS+, together with FOP, is employed as a benchmark method for comparison in the following experiments.

\subsection{Deep-Learning for Learning Samples Distribution}
Learning-based methods augment SBMP by learning a context-conditioned distribution, thereby concentrating samples in feasible and task-relevant regions of the state space.
Early probabilistic learning approaches have been used to bias SBMP via Gaussian Mixture Models (GMMs) learned online from collision and collision-free exemplars, enabling sample biasing toward predicted free regions~\cite{Jinwook2016GMM, Jinwook2017GMM}. However, the method depends on continual acquisition of informative exemplars and careful tuning of parameters.
\textcite{Ma2020ConditionalGA} employ a conditional Generative Adversarial Network (GAN) that operates on rasterized scene representations and predicts dense likelihood maps.
These likelihood maps require additional post-processing to extract planner states and remain tightly coupled with geometric planners, as temporal dynamics are not explicitly modeled.
Transformer-based sampling methods~\cite{Johnson2023Dictionaries, Lei2024TransformerEnhancedMP, Feng2025rrtformer} have also been integrated into SBMP by conditioning sampling proposals on the planning context, but they are typically invoked sequentially at inference time, which scales poorly when large batches of independent samples are required.
Related ideas appear in FOP, where terminal states in the Fren\'{e}t frame are sampled, and then the trajectories are generated by polynomials. LF-Net~\cite{Yu2024LFNet} follows this structure by uniformly discretizing potential terminal states and scoring them with a cross-attention-based classifier. This enables selecting a small set of high-probability candidates before running full trajectory optimization.
Unlike per-sample trajectory evaluation, this reduces computation by shifting most of the candidate scoring process to a parallelizable scoring stage, but it remains tied to discretization since solution quality depends on the range and resolution of the terminal-state lattice.

\textcite{Ichter2018SamplingDistributions} propose a non-uniform sampling strategy for SBMP that favors sampling in those regions where an optimal solution is more likely to lie. A CVAE is used to learn this sampling distribution, conditioned on information specific to each planning problem. The method demonstrates flexibility and effectiveness by learning optimal, conditioned sampling distributions from demonstrations for robotics tasks such as static obstacle avoidance, path planning, and overtaking maneuvers.
The approach presented in this paper builds on this line of work but generalizes it to the domain of autonomous driving. In particular, the BEV-based conditioning allows to handle arbitrary traffic scenarios, rather than being limited to specific maneuvers or predefined subtasks.

\section{Methodology}\label{sec:methodology}
The SP adopts a CVAE model as its core sampling module to learn an efficient trajectory parameters distribution from perception inputs and previous successful plans.
To incorporate temporal dynamics into the CVAE input, three consecutive bird’s-eye-view (BEV) frames are stacked along the channel dimension, generating a conditional image tensor with dimensions $3 \times 256 \times 256$.
Road geometry, obstacle information, and the reference path provided by a global planner are extracted to reconstruct the BEV environment representation.
The ego vehicle is always placed at the center of the image and aligned with a fixed orientation, such that the vehicle’s heading direction coincides with the image’s longitudinal axis, see the BEV image in \autoref{fig:scheme}.
This ego-aligned representation removes appearance variations caused by changing vehicle orientations, thereby reducing input space complexity and improving learning consistency and generalization.
The BEV image is configured as a square region centered on the ego vehicle, covering 50\,m in both longitudinal and lateral directions (i.e., 100\,m $\times$ 100\,m in total).
As the \autoref{alg:cvae} in \autoref{alg:sparse_planner} shows, the CVAE takes a stack of these three images as conditional inputs in its generative configuration, denoted by $y$.
The number of generated samples $N$ is configurable, allowing a flexible trade-off between planning performance and computational cost.
Each sampled trajectory parameter vector $x$ is mapped to a candidate trajectory $\tau_i^{\mathrm{F}}$ in the Fren\'{e}t coordination, which is subsequently transformed into the global coordinate system via the coordinate transformation $\Phi_{\mathrm{F}\rightarrow\mathrm{G}}\!\left(\tau_i^{\mathrm{F}}(t)\right)$.
Subsequent operations such as feasibility checks, collision avoidance, and cost function evaluation remain identical across planners, ensuring a fair comparison between methods.
The cost function $J$ is adopted from \textcite[cf.][Equation (4)]{Sun2023FISSPLUS}.

Since the CVAE does not strictly guarantee that all generated samples satisfy the feasibility constraints, a fallback mechanism is introduced to ensure planning reliability (see \autoref{alg:planB}).
If Sparse Planner fails to produce a feasible trajectory, the classical FOP is invoked to compute a valid solution, which is then returned as the final trajectory.
\autoref{tab:planner_comparison} reports the activation probability of the fallback mechanism in the experiments, which is negligible.

\begin{algorithm}[t]
\caption{Sparse Planner with Parallel Trajectory Generation.}
\label{alg:sparse_planner}
\begin{algorithmic}[1]
\Statex \textbf{Input:} Vehicle-centric BEV conditioning $y$ with lanes, obstacles, and reference path
\Statex \textbf{Output:} Optimal trajectory $\tau^{*}$
\State Sample latent variables from prior \label{alg:cvae}
$z_i \sim p_\xi(z|y)$, sample parameters from decoder $x_i \sim p_\theta(x|z_i, y)$
\; $i = 1,\dots,N$

\For{$i = 1$ \textbf{to} $N$} \label{alg:parallel_loop}
    \State $\tau_i^{\mathrm{F}}(t) = \Psi(x_i)$
    \State $\tau_i^{\mathrm{G}}(t) =
    \Phi_{\mathrm{F}\rightarrow\mathrm{G}}\!\left(\tau_i^{\mathrm{F}}(t)\right)$
    \State $\tau_i^{\mathrm{G}}(t) \in \mathcal{X}_{\mathrm{free}}(t), \;
    g(\tau_i^{\mathrm{G}}(t)) \le 0, \; \forall t \in [0,T]$
    \State $J_i = J\!\left(\tau_i^{\mathrm{G}}\right)$ \label{alg:cost}
\EndFor
\If {$\mathcal{F} = \emptyset$}
    \State \Return FOP\_Algorithm() \label{alg:planB}
\EndIf
\State $i^{*} = \arg\min(J_i), i = 1,\dots,N$
\State \Return $\tau^{*}$
\end{algorithmic}
\end{algorithm}

\subsection{Dataset Recording}
\label{subsec:dataset}
The experiments utilize traffic scenarios from the CommonRoad\footnote{https://commonroad.in.tum.de/scenarios} framework.
To generate supervision signals for training the CVAE, dense uniform sampling is first performed in the Fren\'{e}t parameter space $(d, \dot{s}, T)$ using the FOP.
The optimal trajectory selected by FOP is then used to extract the corresponding terminal sampling parameters, which serve as the learning targets for the CVAE model.
In the experiments, the number of samples, sampling ranges and resolution are defined as
$d \in [-1.7, 1.7]\,\mathrm{m}$ (34 samples, 0.1\,m resolution),
$\dot{s} \in [0, 14]\,\mathrm{m/s}$ (28 samples, 0.5\,m/s resolution),
and $T \in [3, 5]\,\mathrm{s}$ (20 samples, 0.1\,s resolution).
The total number of samples per planning cycle is 19040.
Experiments are conducted on 3500 scenarios, each lasting 5–15\,s, randomly divided into training, validation, and test sets with a 70\%/20\%/10\% split. The 10\% test split (350 scenarios) is used for planner-level evaluation.

Since CommonRoad scenarios are provided at a temporal resolution of $0.1\,\mathrm{s}$,
BEV sequences are built at the same frequency to ensure temporal consistency.
The first input contains three identical initial frames; the second contains two initial frames and one subsequent frame.
From the third step onward, three consecutive frames are used, forming a sliding window.
Each stacked BEV input is paired with its associated optimal sampling parameters,
forming input--output training pairs for the CVAE model.
The resulting dataset is used for model training and validation.

\begin{figure*}[t]
    \centering
    \includegraphics[width=0.9\linewidth]{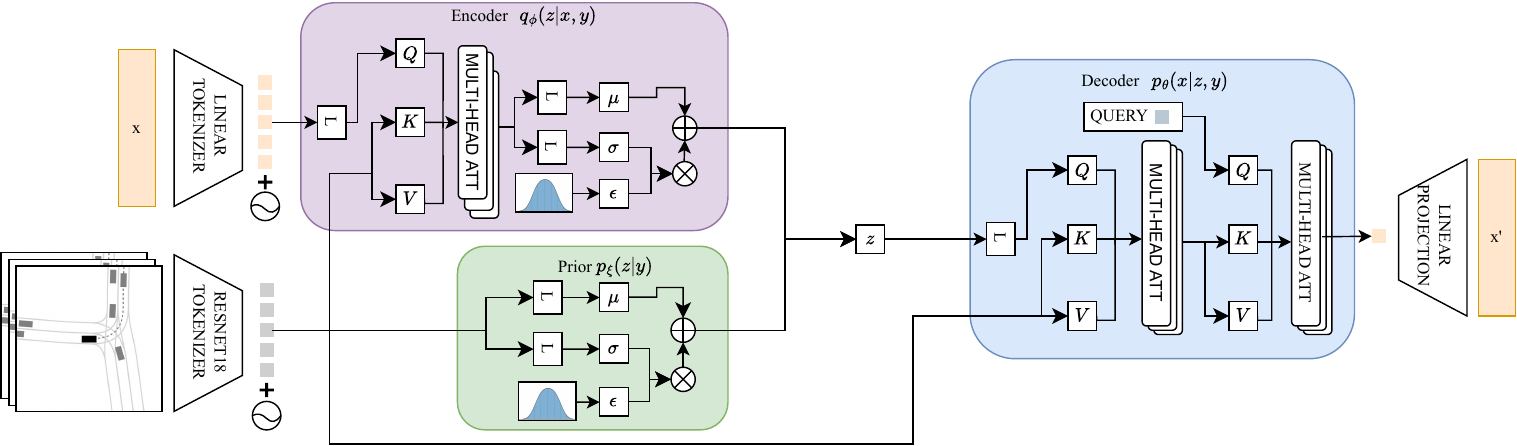}
    \caption{Architecture of the CVAE. During training, the encoder–decoder modules map the distribution of the optimal parameters, conditioned on the BEV embeddings, into a latent space and back to the parameter space. During inference, the encoder is replaced by the prior, from which the latent representation is sampled, while the same decoder as used during training generates the output.}
    \label{fig:architectures}
\end{figure*}

\subsection{CVAE Training}
The scope of the model is to learn distributions in the parameter space containing optimal trajectories and conditioned on the context, in this case the BEV sequence of the traffic scenario. Let $x$ denote a sample in the parameter space, $y$ the encoding of the conditional context, and $z$ a sample in the latent space. The structure of a CVAE consists in two neural networks, the probabilistic encoder $q_{\phi}(z | x, y)$, with weights of the network denoted by $\phi$, and the probabilistic decoder $p_{\theta}(x |  z, y)$, with weights $\theta$. The probabilistic encoder approximates the posterior latent distribution $p(z | x, y)$, conditioned on the context $y$ and on the original sample $x$, the probabilistic decoder models the likelihood of the sample $x$ conditioned on the latent representation $z$ and on the same context $y$.

The encoder maps the sample $x$ into a multivariate normal latent distribution through a neural network $q_{\phi}(z | x, y) = \mathcal{N}(z |\mu_{\phi}(x,y), \sigma^{2}_{\phi}(x, y) I)$, where the mean $\mu_{\phi}(x,y)$ and the variance $\sigma^{2}_{\phi}(x, y)$ of the approximate posterior are the outputs of the encoding neural network. The reparameterization trick $z = \mu_{\phi}(x,y) + \sigma_{\phi}(x, y) \odot \epsilon, \ \ \epsilon \sim \mathcal{N}(0, I)$ allows sampling from the approximate posterior in a differentiable way, enabling gradients to be computed with respect to the encoder parameters.

The generative model defines the joint distribution $p(x, z | y) = p_{\theta}(x | z, y) p(z|y)$, which factorizes into the product of the probabilistic decoder (the likelihood term) and the prior distribution over the latent variables $p(z|y)$. The conditioned prior is defined as $p_{\xi}(z|y) = \mathcal{N}(z | \mu_{\xi}(y), \sigma^{2}_{\xi}(y) I )$, where the mean $\mu_{\xi}(y)$ and variance $\sigma^{2}_{\xi}( y)$ are the outputs of a linear projection with weights $\xi$. The probabilistic decoder is modeled as $p_{\theta}(x | z, y) = \mathcal{N}(x | f_{\theta}(z, y), \sigma^2I)$, where $\sigma^2$ is set to be a small value and $f$ is a deterministic function encoded as a neural network with weights $\theta$ that models the mean of the distribution.

During training, the parameters $\theta$, $\phi$, and $\xi$ are jointly optimized to reduce the reconstruction error between the input $x$ and the decoder output $f_{\theta}(z, y)$, and to reduce the difference between the approximated posterior $q_{\phi}(z | x, y)$ and the true posterior $p(z | x, y)$. To achieve this, the evidence lower bound (ELBO) is maximized \cite{kingma2013VAETheory}:
\begin{equation}
\mathcal{L}_{\theta, \phi, \xi}(x|y) = \log{p_{\theta}(x|y)} - D_{KL}(q_{\phi}(z | x, y)||p(z | x, y))
\label{eq:ELBO}
\end{equation}
that corresponds to maximize the log-likelihood of the observed data (conditioned on the context $y$) and to minimize the Kullback–Leibler divergence of the approximate from the true posterior. \autoref{eq:ELBO} must be reformulated accordingly because the terms $p_{\theta}(x|y)$ and $p(z | x, y)$ are intractable. Considering that $p_{\theta}(x|z,y) = \mathcal{N}(x|f_{\theta}(z,y), \sigma^2I)$, the ELBO is, up to an additive constant, proportional to:
\begin{equation}
\begin{aligned}
\mathcal{L}_{\theta, \phi, \xi}(x|y) \propto - \mathbb{E}_{z \sim q_{\phi}(z|x,y)}[||x - f_{\theta}(z,y)||_{2}^2] \ + \\ - \ D_{KL}(q_{\phi}(z | x, y)||p_{\xi}(z|y))
\end{aligned}
\label{eq:ELBO3}
\end{equation}
Where the Kullback-Leibler divergence is reformulated using the prior $p_{\xi}(z|y)$.
The CVAE is then trained using the following loss:
\begin{equation}
\begin{aligned}
\theta^{*}, \phi^{*}, \xi^{*} = \arg \min_{\theta , \phi, \xi} - \mathcal{L}_{\theta, \phi, \xi}(x|y) \ + \\
+ \ \mathbb{E}_{z \sim p_{\xi}(z|y)}[||x - f_{\theta}(z,y)||_{2}^2]
\end{aligned}
\label{eq:loss}
\end{equation}
Where the first term is the negative ELBO, while the second term is the mean squared error (MSE) loss between the generated sample and the ground-truth data.

\subsection{CVAE Architecture}
\label{sec:CVAE_optimization}
The architecture for training and generation is illustrated in \autoref{fig:architectures}.
In the following, the main components of the architecture are analyzed.

\textbf{Tokenization}: The three conditioning frames $y$ are stack into a single tensor of shape (3, 256, 256). After normalization, this tensor is then processed through a pretrained ResNet18 model to produce a sequence of tokens of shape (64, 32). The network input $x = (d, \dot{s}, T)$ is also tokenized into a tensor of shape (16, 32). Learned positional encoding vectors are added to both token sequences.

\textbf{Encoder}: The tokenized input is fed as a query to a multi-head cross-attention layer, which forms the backbone of the encoder, with keys and values coming from the BEV context. The resulting output is pooled along the token sequence dimension and linearly projected to produce the parameters of the latent space distribution, $\mu$ and $\log \sigma^2$, but tensors of shape (64). The latent representation $z$ is then sampled from this distribution through the reparameterization trick.

\textbf{Decoder}: The sampled latent representation $z$ is projected into the token space via a linear tokenizer. This (16, 32) query then attends to the conditioning representation $y$ through cross-attention. Finally, a learned (1, 32) query vector attends over the token sequence via cross-attention, aggregating the information into a single representation. The output is projected back into the input space.

\textbf{Prior}: During conditioned generation, the posterior distribution modeled by the encoder is replaced with the prior distribution, whose parameters $\mu$ and $\log \sigma^2$ are computed via two linear layers. In this work, the prior is conditioned on the BEV context, $p(z|y)$.

The CVAE proposed in this work features novel aspects in its architectural design and training strategy. Specifically:
\begin{itemize}
    \item In standard CVAEs, the loss function typically corresponds to the negative ELBO. In this work, the loss is augmented with an additional generative MSE term (see \autoref{eq:loss}).
    \item The prior in the CVAE is not modeled as a standard normal distribution, as is common in the literature, but is instead conditioned on the context. In standard CVAEs, the conditioning is applied only in the decoder.
\end{itemize}

Experiments show that incorporating the loss from \autoref{eq:loss} reduces MSE by about 14\%, while using a context-conditioned prior yields an additional 10\% reduction.

\section{Evaluation}
\label{sec:evaluation}
As described in \autoref{subsec:dataset}, 350 scenarios are used to evaluate the planners. For each scenario, the runtime per planning cycle is recorded during simulation, and after completing the full trajectory, the total trajectory cost is computed.
It is important to note that this experiment assumes the input images to the CVAE are provided by a preceding prediction module, and the runtime of this module is not included in the reported runtime of SP.
After all scenarios are executed, the statistics are aggregated across the entire test set. The average trajectory cost $\bar{J}$ and the average runtime per planning cycle $\bar{t}$ are calculated over all scenarios. For SP, whenever no feasible trajectory is found in a planning cycle, the planner falls back to FOP. The number of such fallback events is accumulated across all scenarios and normalized by the total number of planning cycles to obtain the average fallback rate $p_{\mathrm{B}}$, which reflects the reliability of the CVAE-based sampling.
The number of samples used by FOP in the fallback mechanism is set to 125, i.e., 5 samples for each variable $d$, $\dot{s}$, and $T$.

The trajectory cost is computed using the same cost function as FISS+, as defined in \textcite[cf.][Equation (7)]{Sun2023FISSPLUS}:
\begin{equation}
\label{eq:cost_function}
\begin{aligned}
J_{\mathrm{total}} &=
w_V \int_{t_0}^{t_f} (v_{\mathrm{target}}-v)^2 \, dt
+ w_A \int_{t_0}^{t_f} a^2 \, dt\\
&\quad
+ w_J \int_{t_0}^{t_f} \dot{a}^2 \, dt
+ w_D \int_{t_0}^{t_f} \max(\xi_1,\ldots,\xi_o)\, dt\\
&\quad
+ w_{LC} \int_{t_0}^{t_f} d^2\, dt
+ w_T t_f .
\end{aligned}
\end{equation}
The parameter $v_{\mathrm{target}}$ denotes the desired velocity, $a$ and $\dot{a}$ represent the acceleration and jerk, and $t_f$ denotes the terminal time.
The term $\xi_i = e^{-w_{dist}d_i}$ penalizes proximity to obstacles, where $d_i$ is the distance between the ego vehicle and the $i$-th obstacle, and $o$ denotes the total number of surrounding obstacles. The term $d$ is the lateral deviation of the ego vehicle from the reference path (lane center) at time step $t$.
The weights are set as follows: $w_V=0.1$, $w_A=10$, $w_J=10$, $w_D=100$, $w_{dist}=0.1$, $w_T=1$, $w_{LC}=10$.
This set of weights prioritizes ride comfort and trajectory safety, emphasizing acceleration, jerk, obstacle proximity, and lateral deviation from the lane center.

The experiments are conducted on a system equipped with an Intel Core i7-12700H CPU, and the CVAE inference is accelerated using a NVIDIA GeForce RTX 3050 GPU.

\begin{figure*}[t]
    \centering
    \begin{subfigure}[t]{0.48\textwidth}
        \centering
        \includegraphics[height=4.4cm]{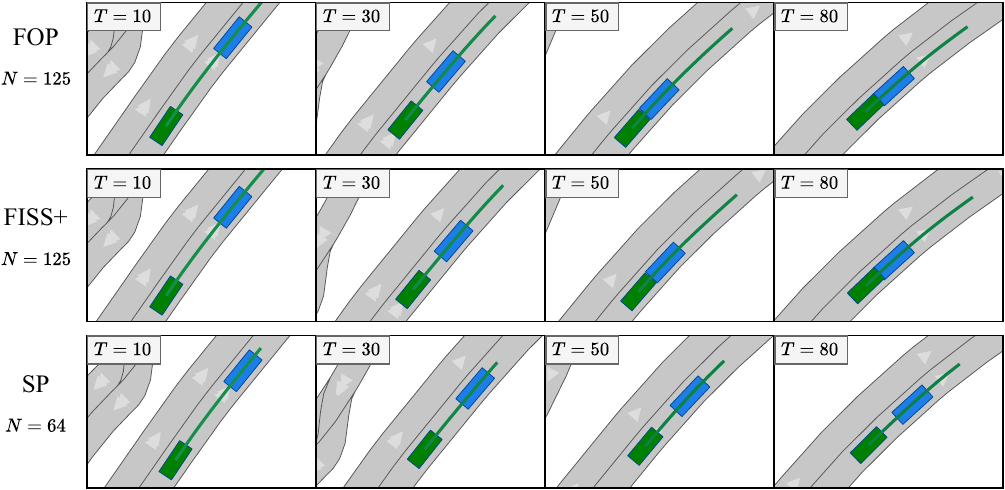}
        \caption{Scenario: USA\_NewYork-54\_1\_T-1.}
        \label{fig:newyork}
    \end{subfigure}
    \hfill
    \begin{subfigure}[t]{0.48\textwidth}
        \centering
        \includegraphics[height=4.4cm]{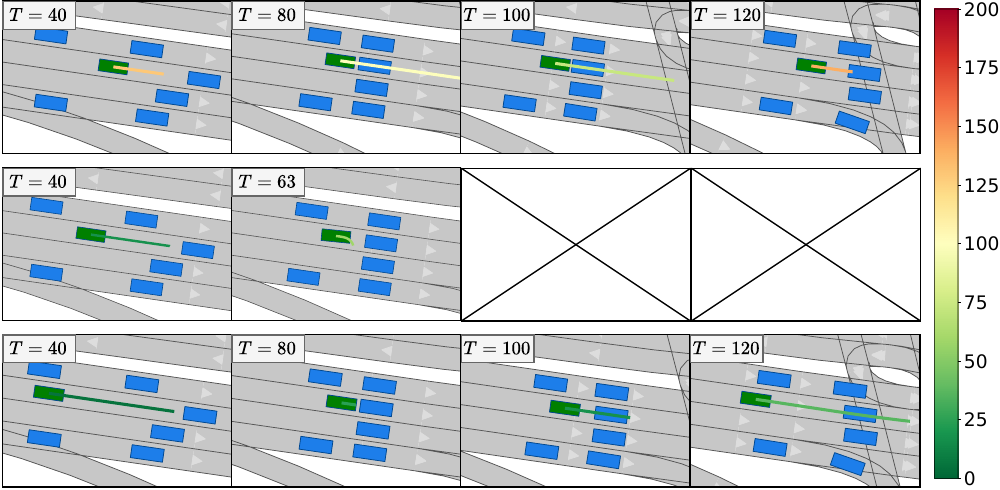}
        \caption{Scenario: CHN\_Qingdao-5\_37\_T-1.}
        \label{fig:qingdao}
    \end{subfigure}

    \caption{Qualitative comparison of the planners in two representative scenarios.}
    \label{fig:scenario_brake}
\end{figure*}

\begin{table}[t]
\caption{Performance comparison under sample number  $N$.}
\label{tab:planner_comparison}
\centering
\scriptsize
\setlength{\tabcolsep}{3pt}
\renewcommand{\arraystretch}{1.1}

\begin{tabular}{llcccccccc}
\toprule
& & \multicolumn{8}{c}{$N$} \\
\cmidrule(lr){3-10}
Metric & Method & 1 & 8 & 16 & \textbf{64} & 125 & 216 & 512 & 1000 \\
\midrule

\multirow{3}{*}{$\bar{J}$}
& FOP   & -- & -- & -- & -- & 74.81 & 72.94 & 71.60 & 71.13 \\
& FISS+ & -- & -- & -- & -- & 72.75 & 71.90 & 71.19 & 71.01 \\
& SP    & 72.50 & 71.66 & 71.35 & \textbf{71.16} & 70.85 & 70.67 & 70.57 & 70.48 \\
\midrule
\multirow{3}{*}{$\bar{t}$ [ms]}
& FOP   & -- & -- & -- & -- & 72 & 124 & 320 & 596 \\
& FISS+ & -- & -- & -- & -- & 57 & 70 & 115 & 240 \\
& SP    & 11 & 15 & 33 & \textbf{47} & 82 & 163.4 & 330 & 669 \\
\midrule
\multirow{1}{*}{$p_{B}$ (\%)}
& SP    & 18.9 & 10.1 & 5.3 & \textbf{4.0} & 4.3 & 4.4 & 4.4 & 4.9 \\
\midrule
\multirow{1}{*}{$N_{fail}$ (\%)}
& FISS+    & - & - & - & - & 23.4 & 20.5 & 20.0 & 20.0 \\

\bottomrule
\end{tabular}
\end{table}

\subsection{Performance Analysis}
The experiments evaluate FOP, FISS+, and SP on 350 scenarios.
Since FISS+ fails to generate a feasible trajectory in 82 cases, the performance comparison is conducted on the remaining 268 scenarios where all planners successfully reach the goal.
In contrast, FOP and SP---with FOP as a fallback---successfully complete all 350 scenarios.
The percentage of scenarios in which FISS+ fails to reach the goal is denoted by $N_{fail}$ and summarized in the performance comparison table (see \autoref{tab:planner_comparison}).
When the sampling number increases from $N=125$ to $N=216$, the $N_{fail}$ decreases from 23.4\% to 20.5\%, but remains around 20\% thereafter. This suggests that the algorithm exhibits limitations in handling certain challenging scenarios. Representative examples are discussed in \autoref{subsec:case_study}.
For FOP and FISS+, metrics are reported starting from $N = 125$, because a smaller sample number does not consistently produce feasible trajectories across all scenarios.
Furthermore, $N = 216$, $512$, and $1000$ correspond to 6, 8, and 10 samples per variable, respectively.

\begin{figure}[t]
    \centering
    \includegraphics[width=\linewidth]{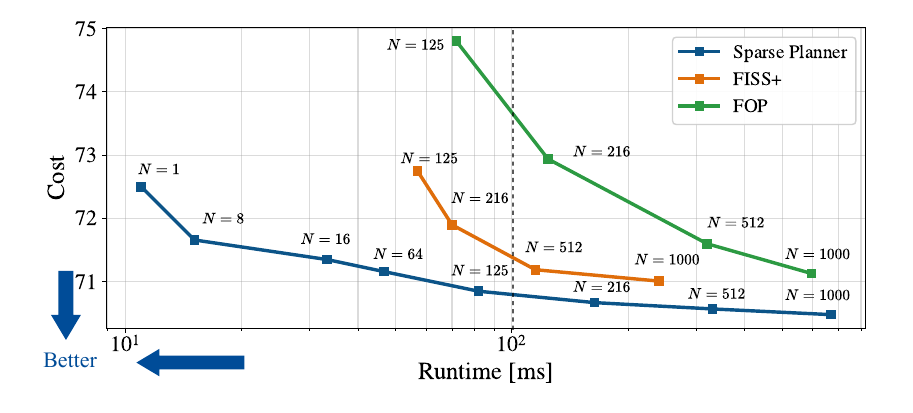}
    \caption{Performance visualization based on \autoref{tab:planner_comparison}.}
    \label{fig:efficientVScost}
\end{figure}

Across all planners, increasing the sampling number $N$ generally improves reducing trajectory cost at the expense of higher runtime.
For SP and FOP, runtime scales approximately linearly with $N$, reflecting the per-sample trajectory evaluation complexity.
In contrast, FISS+ exhibits significantly slower runtime growth as $N$ increases, as illustrated in \autoref{fig:efficientVScost}. This behavior results from its heuristic prioritization strategy, which avoids evaluating all samples.

SP consistently achieve a lower trajectory cost than FOP and FISS+ at comparable sampling sizes. Notably, SP with only $N=1$ already outperforms both FOP and FISS+ at $N=125$ in terms of trajectory cost.
This result indicates that the CVAE model effectively captures the relationship between environmental context and low-cost trajectory parameters, and tends to generate samples that implicitly encode obstacle-avoidance behavior.

However, SP shows a relatively high fallback rate at $N=1$ (18.9\%). As $N$ increases, the fallback rate decreases significantly, converging to 4.0\% at $N=64$. Beyond $N=64$, further increasing $N$ does not reduce the fallback rate, suggesting that this remaining percentage corresponds to long-tail scenarios the model fails to map to a correct optimal-parameter distribution.

The runtime of SP nearly doubles from $N=64$ (47\,ms) to $N=125$ (82\,ms), while the improvement in trajectory cost is marginal and the fallback rate has already stabilized.
Therefore, $N=64$ is selected as the optimal sampling number for SP, providing the best trade-off between solution quality, computational efficiency, and reliability. Moreover, the average inference time of the CVAE model for generating 64 samples is 7\,ms per planning cycle, which introduces only minor overhead.

\subsection{Qualitative Case Study}
\label{subsec:case_study}
To further examine the behavioral differences under challenging traffic conditions, \autoref{fig:scenario_brake} presents two representative scenarios that illustrate the advantages of the proposed planner over the original FOP and FISS+.

The scenario in \autoref{fig:newyork} represents a lane following scenario with a dynamic obstacle in front of the ego vehicle. Remarkably, SP plans the safest trajectories at time steps 50 and 80.
SP exhibits increased longitudinal clearance from the obstacle in front, reducing the tailgating risk.
This behavior stems from the ability of SP to leverage the CVAE model trained on~2450 scenarios planned by FOP with a large number of samples, as explained in \autoref{subsec:dataset}.
As a result, SP can represent high-quality trajectories that would be infeasible to compute in real time with FOP using highly dense sampling.

The scenario in \autoref{fig:qingdao} represents a drive-brake-drive maneuver which challenges the assumption of FISS+ heuristic that the next sample is similar to the last one.
At time step 63, FISS+ fails to plan a feasible trajectory because it selects trajectories close to the previous solution with minimal cost, leading to insufficient deceleration. When the leading vehicle stops, the ego vehicle approaches too closely, leaving insufficient braking distance and causing planning failure.
In contrast, FOP and SP successfully plan the full scenario, with SP achieving a lower trajectory cost and maintaining a greater distance from the front vehicle, particularly at time step 100.

\subsection{Deterministic Runtime Performance}
\begin{figure}[t]
    \centering
    \includegraphics[width=1.0\linewidth]{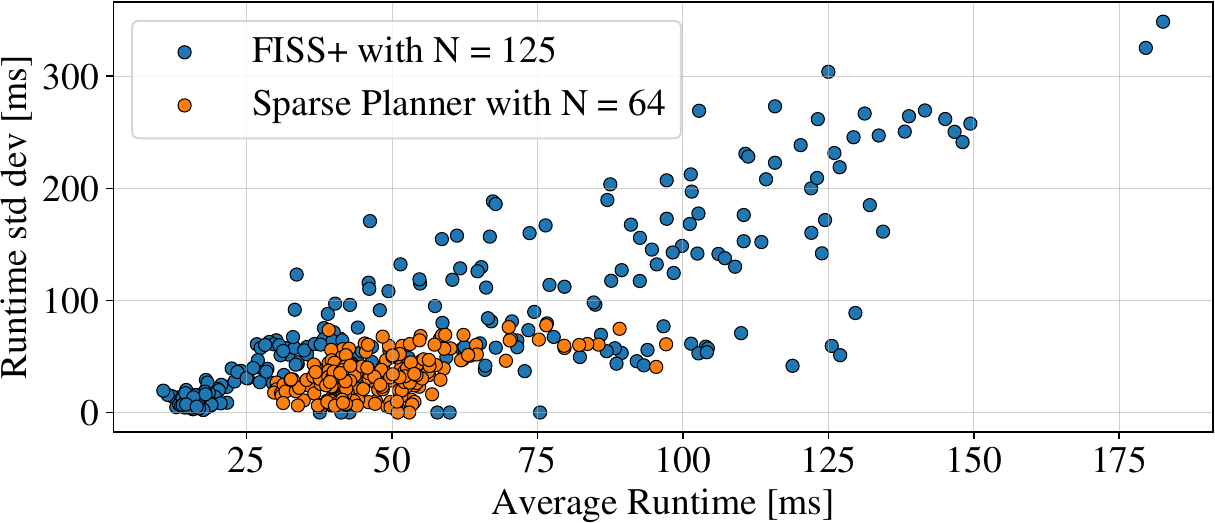}
    \caption{Runtime variability across 268 scenarios for FISS+ and SP. Each point represents a scenario, plotted by its average runtime and corresponding standard deviation across scenario planning cycles.}
\label{fig:runtime_variability}
\end{figure}

The runtime characteristics of FISS+ and SP, shown in \autoref{fig:runtime_variability}, reveal a fundamental distinction in deterministic runtime performance between the two planning methods. FISS+ exhibits a wide spread in both mean runtime and runtime standard deviation across scenarios, with points scattered broadly across the plot.
This behavior arises from its reliance on heuristic search strategies, whose performance can degrade unpredictably depending on the scenario.
In the worst-case scenarios, two FISS+ instances reach a mean runtime of over 175\,ms with a standard deviation exceeding over 300\,ms.
This variability makes FISS+ poorly suited for safety-critical applications, where deterministic runtime guarantees are essential.
In contrast, SP produces a compact, low-variance distribution clustered in the lower-left region of the plot, reflecting more deterministic runtime performance, making it more suitable for real-time system such as autonomous driving system.

\section{Conclusion}
This work presents the Sparse Planner, a hybrid planning framework that leverages a CVAE-based sampling strategy to efficiently sample from the parameter space. By learning a context-conditioned sampling distribution, SP generates low-cost trajectory parameters with significantly fewer samples, thereby substantially reducing runtime.

The learned sampling mechanism exhibits strong performance, particularly in scenarios with dynamic obstacles, where contextual understanding is crucial. Compared to the state of the art of sampling-based planners, SP achieves lower trajectory cost while maintaining low, bounded, and deterministic runtime behavior. SP also offers inherent potential for parallelization, providing further opportunities to enhance real-time performance.

The BEV representation, used as a conditioning context, enables compatibility with real sensor setups, while the fallback mechanism ensures completeness in case of failure, making SP highly suitable for real-time autonomous driving systems. Future work could explore integrating SP with a real sensor stack, moving the approach closer to a sensor-driven pipeline.

\printbibliography

\end{document}